\documentclass[sigconf]{acmart}

\usepackage{graphicx} 
\usepackage{subfigure} 
\usepackage{amsmath} 
\usepackage{balance} 

\copyrightyear{2024}
\acmYear{2024}
\setcopyright{acmlicensed}\acmConference[WSDM '24]{Proceedings of the 17th ACM International Conference on Web Search and Data Mining}{March 4--8, 2024}{Merida, Mexico}
\acmBooktitle{Proceedings of the 17th ACM International Conference on Web Search and Data Mining (WSDM '24), March 4--8, 2024, Merida, Mexico}
\acmPrice{15.00}
\acmDOI{10.1145/3616855.3635783}
\acmISBN{979-8-4007-0371-3/24/03}

\acmSubmissionID{wsdmfp218}

\begin{document}

\title{PhoGAD: Graph-based Anomaly Behavior Detection with Persistent Homology Optimization}

\author{Ziqi Yuan}
\orcid{0000-0001-8539-3146}
\affiliation{
	\department{School of Computer Science and Engineering}
	\institution{Beihang University}
	\city{Beijing}
	\country{China}
	\postcode{100191}
}
\email{yuanzq@buaa.edu.cn}

\author{Haoyi Zhou}
\orcid{0000-0002-2393-3634}
\affiliation{
	\institution{Zhongguancun Laboratory}
	\city{Beijing}
	\country{China}
	\postcode{100191}
}
\affiliation{
	\department{School of Software}
	\institution{Beihang University}
	\city{Beijing}
	\country{China}
	\postcode{100191}
}
\email{haoyi@buaa.edu.cn}

\author{Tianyu Chen}
\orcid{0009-0002-0649-5799}
\affiliation{
	\department{School of Computer Science and Engineering}
	\institution{Beihang University}
	\city{Beijing}
	\country{China}
	\postcode{100191}
}
\email{tianyuc@buaa.edu.cn}

\author{Jianxin Li}
\authornote{Corresponding author.}
\orcid{0000-0001-5152-0055}
\affiliation{
	\institution{Zhongguancun Laboratory}
	\city{Beijing}
	\country{China}
	\postcode{100191}
}
\affiliation{
	\department{School of Computer Science and Engineering}
	\institution{Beihang University}
	\city{Beijing}
	\country{China}
	\postcode{100191}
}
\email{lijx@buaa.edu.cn}

\renewcommand{\shortauthors}{Ziqi Yuan, Haoyi Zhou, Tianyu Chen, and Jianxin Li}

\begin{abstract}

A multitude of toxic online behaviors, ranging from network attacks to anonymous traffic and spam, have severely disrupted the smooth operation of networks. Due to the inherent sender-receiver nature of network behaviors, graph-based frameworks are commonly used for detecting anomalous behaviors. However, in real-world scenarios, the boundary between normal and anomalous behaviors tends to be ambiguous. The local heterophily of graphs interferes with the detection, and existing methods based on nodes or edges introduce unwanted noise into representation results, thereby impacting the effectiveness of detection.
To address these issues, we propose PhoGAD, a graph-based anomaly detection framework. PhoGAD leverages persistent homology optimization to clarify behavioral boundaries. Building upon this, the weights of adjacent edges are designed to mitigate the effects of local heterophily. Subsequently, to tackle the noise problem, we conduct a formal analysis and propose a disentangled representation-based explicit embedding method, ultimately achieving anomaly behavior detection.
Experiments on intrusion, traffic, and spam datasets verify that PhoGAD has surpassed the performance of state-of-the-art (SOTA) frameworks in detection efficacy. Notably, PhoGAD demonstrates robust detection even with diminished anomaly proportions, highlighting its applicability to real-world scenarios. The analysis of persistent homology demonstrates its effectiveness in capturing the topological structure formed by normal edge features. Additionally, ablation experiments validate the effectiveness of the innovative mechanisms integrated within PhoGAD.

\end{abstract}



\maketitle

\section{Introduction}

Nowadays, with the continuous expansion of the global Internet, the volume of data and the complexity of online behaviors have increased significantly. Various forms of toxic online behaviors, such as network attacks, anonymous traffic, and spam emails, have disrupted the normal use of networks. Consequently, the network anomaly behavior detection field has attracted increasing attention.

Network behaviors, being inherently temporal data, have been extensively studied for modeling normal behavior sequences to detect anomalous behaviors. Building upon this foundation and leveraging the inherent nature of network behavior with initiators and targets, graph structures adept at expressing data correlations are introduced into this domain for data representation. Graph models utilize nodes to represent entities such as users and IP addresses, and edges to represent specific behaviors. A series of methods based on node and edge detection utilize graph models such as attack graphs and provenance graphs to achieve more precise anomaly behavior detection.

However, in real-world scenarios, the initiators of anomalous behaviors often adjust their behavioral patterns to evade detection. For instance, attackers mimic the frequency of normal accesses, and spam emails emulate the content of regular emails. This blurs the boundaries between normal and anomalous behavior in the corresponding graph structure, where anomalous edges are no longer concentrated but instead are adjacent to normal edges. This results in local heterophily within graphs.

\begin{figure}[h]
	\centering
	\subfigure[Node-Based Representation]{\includegraphics[width=0.48\linewidth]{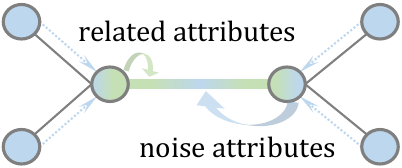}}
	\hfill
	\subfigure[Edge Convolution]{\includegraphics[width=0.48\linewidth]{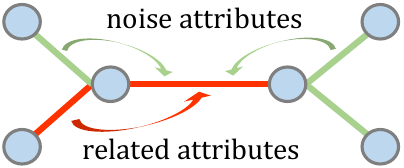}}
	
	\caption{The noise problem presented by current detection methods. (a) demonstrates the noise problem arising from node-based edge representation, where blue and green represent nodes and their attributes directly and indirectly linked to the edge. (b) showcases the noise problem arising from edge convolution, where green and red signify normal and anomalous edges.}
	
\Description{The challenges posed by existing detection methods are illustrated through two subfigures. (a) illustrates the noise problem caused by node-based edge representation methods, particularly the introduction of unrelated node attributes into the edges. (b) demonstrates the noise problem caused by edge convolution methods, specifically the attribute confusion introduced by adjacent edges of different classes.}

\label{fig_issues_schematic}
\end{figure}

Existing node-based embedding methods for edge representation introduce attributes of nodes not directly connected to the target edge, forming noise from unrelated node attributes. On the other hand, methods that directly perform convolution on edges face challenges in adapting to the local heterophily. During the convolution, representations of anomalous edges are influenced by the adjacent normal edge attributes, also resulting in the noise.
Figure \ref{fig_issues_schematic} illustrates the introduction of the noise problem. This problem results in insufficient distinguishability of behavioral representations and, consequently, limiting the effectiveness of anomaly behavior detection.

To address this issue, we conduct a formal analysis of the noise present in current methods and introduce PhoGAD, a graph-based anomaly behavior detection framework with persistent homology optimization. Specifically, we utilize persistent homology to analyze enduring structures in the feature space, treating them as normal signals, and encouraging attribute similarity among related edges to enhance differentiation between attributes of anomalous and normal edges. We devise a disentangled representation mechanism and calculate neighbor weights based on second-order neighboring nodes, adapting to the local heterophily and preventing the introduction of noise. These mechanisms empower PhoGAD to explicitly embed edges, acquiring distinctive edge representations that enhance the direct detection of edges corresponding to specific behaviors in the output layer.

Our contributions could be summarized as follows:

\begin{itemize}

\item We point out the local heterophily within the graph structures faced by existing methods and conduct a formal analysis of the noise problem it triggers.

\item We design an edge explicit embedding method with adjacency edge weights and disentangled representation to accommodate the local heterophily, enhancing data adaptability and resolving the noise problem.

\item We utilize a topological approach based on persistent homology to optimize edge attributes, mitigating the impact of blurred behavioral boundaries on detection.

\item The proposed PhoGAD framework demonstrates remarkable generalization, surpassing SOTA methods across diverse scenarios and adapting effectively to extremely low anomaly proportions.

\end{itemize}

The paper is organized as follows: Section \ref{sec_related_work} introduces relevant research related to this study. Section \ref{sec_problem_formulation} formally presents the practical challenges encountered by anomaly behavior detection methods. In Section \ref{sec_framework}, the PhoGAD framework proposed in this paper is introduced. Section \ref{sec_experiments} presents the experimental setup, along with the results and analysis. Lastly, Section \ref{sec_conclusion} concludes the paper.

\section{Related Work}
\label{sec_related_work}

In the early stages of research on network anomaly behavior detection, many methods are based on features and sequence relationships. However, with the progression of graph deep learning and other technological advancements, graph-based detection methods have taken the forefront.

\subsection{Detection based on features and sequences}

The emergence and evolution of machine learning and deep learning has facilitated the utilization of feature-based detection methods, such as \cite{APE_add_ref}. However, these methods frequently require the manual delineation of behavioral features and demonstrate a significant dependence on domain expertise. Subsequently, sequence-based techniques were introduced, with DeepLog \cite{DeepLog} standing as a notable example. These methods model sequences of normal behaviors, identifying behaviors that diverge from established sequence patterns as anomalies.

LSTM \cite{LSTM}, Transformer \cite{transformer}, and other models help researchers delve into deeper sequence features for anomaly detection. Methods like LayerLog \cite{LayerLog}, CAT \cite{CAT}, and BPESequence \cite{BPESequence} have emphasized the exploration of multi-level semantic correlations, spanning from individual words to sequences, intending to acquire more precise representations of behaviors. On the other hand, RADT \cite{RADT} and MADDC \cite{MADDC} have focused on contextual information to enhance detection performance and robustness.

\subsection{Detection based on Graphs}

Benefiting from the advancements in graph learning \cite{new_gl2, new_gl3, GCN+}, exemplified by GCN \cite{GCN} and DGI \cite{DGI}, a series of graph-based detection methods have been applied to behavior detection, enhancing the distinguishability of representation results \cite{SUGAR}. These detection methods can be broadly classified into node-based detection and edge-based detection.

\cite{standard_node_detection} represents the most typical node-based detection method, involving the embedding of nodes into a lower-dimensional space, followed by a binary classifier for anomaly detection. Expanding upon this foundational framework, methods like CoLA \cite{CoLA} and LogLG \cite{LogLG} introduce novel indicators, such as the similarity between positive and negative sample pairs, to identify anomalies. Moreover, based on various graph autoencoders \cite{gae1, gae2}, methods like AdONE \cite{AdONE} and AnomalyDAE \cite{AnomalyDAE} compute the reconstruction errors of nodes, thereby deriving anomaly scores.

The research specifically focused on edge detection is relatively limited in comparison to node detection. Early studies directly detected anomalous edges based on the graph structure. However, many recent studies have adopted implicit representation models for edges, where the edge representations are computed based on corresponding node representations. Noteworthy methods within this category comprise CSAWD \cite{CSAWD}, UGED \cite{UGED}, and AANE \cite{AANE}.

Nevertheless, studies such as AddGraph \cite{AddGraph} and ICANE \cite{ICANE} have pointed out the drawbacks of implicit representation and proposed explicit embedding models for edges. These models adopt sequential information, edge attributes, and other factors to optimize the detection results. GNN-NIDS \cite{GNN-NIDS} and other methods based on this approach have found applications in specific domains. However, effective explicit embedding methods for edges are still an area requiring further research and exploration.

\section{Preliminaries and Problem Formulation}
\label{sec_problem_formulation}

\subsection{Anomaly Behaviors}
\label{sec_anomaly_behaviors}
Anomaly network behavior encompasses a wide-ranging concept, and the exact definition of what constitutes an anomaly remains elusive \cite{ad_survey_gan}. As a result, tailored detection methods have been continuously proposed to address distinct anomalous behaviors \cite{provenance_graph_ids_survey, spam_survey}. These methods often use domain-specific prior knowledge to attain favorable detection results.

In our approach, we adopt a unified framework concentrating on three primary categories of network anomalous behaviors.

\begin{itemize}
	\item \textit{Network intrusion.} 
	Traditional attack methods often adhere to conventional paths, yet attackers frequently veil their behaviors, posing obstacles to detection. Detecting such behaviors significantly contributes to upholding network and data security.
	
	\item \textit{Anonymous traffic.} 
	While anonymous traffic represented by TOR (The Onion Router) can be utilized for legitimate intentions, it is frequently used for hiding malicious activities such as online fraud. Detecting Tor traffic enables effective prevention of potential malicious behaviors and ensures compliance in network usage.
	
	\item \textit{Spam emails.}: 
	Spam emails frequently contain advertisements or deceptive information, leading to annoyance and imposing security hazards on users. Detecting spam emails helps protect network bandwidth from occupation and safeguards users' experience and productivity.
\end{itemize}

Although these behaviors exhibit distinct anomaly characteristics, they collectively share the fundamental attribute of accesses. Consequently, we approach the detection with an access-oriented perspective, employing graphs to accomplish anomaly behavior detection through edge detection.

\subsection{Local Heterophily}

In the context of graph anomaly detection, the proportion of anomalies is notably low, resulting in a generally low level of graph heterophily. This implies that connected nodes or edges mostly share the same labels. However, within the local structures where anomalous edges exist, the degree of heterophily experiences a sharp escalation.

Certain research in graph learning is dedicated to addressing the challenges posed by heterophilous graphs \cite{heterophilous_graph}. Traditional research mainly concentrates on scenarios where different nodes are connected, with relatively limited attention given to exploring the heterophily inherent in edges.

In the context of anomaly detection, conventional homophilous graph learning methods potentially lead to representation convergence, resulting in the overshadowing of anomalous embeddings by normal features. Conversely, heterophilous graph learning methods exhibit suboptimal performance when confronted with homophilous structures, making them less effective for anomaly detection.

\subsection{The Noise Problem}

Edge-based detection methods are widely recognized as the most accurate category of anomaly behavior detection methods. However, existing methods frequently compute edge representations through node embeddings, thus introducing irrelevant attribute noise into the edge representation and consequently affecting the quality of the representations.

\begin{figure}[h]
	\centering
	\includegraphics[width=\linewidth]{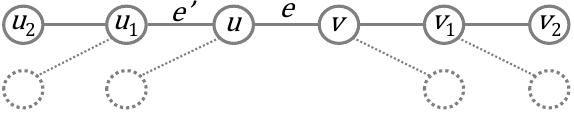}
	\caption{An illustrative figure of a graph structure. Solid lines represent nodes and edges of interest, while dashed lines represent irrelevant portions that have no bearing on the illustration. Disregarding the direction of edges will not hinder the analysis of the noise problems.}
	\Description{An illustrative figure of a graph structure. The figure reveals six interconnected nodes remaining after excluding irrelevant structures represented by dashed lines. From left to right, these nodes are labeled as u2, u1, u, v, v1, and v2. The edge connecting u1 and u is labeled as e1, while the edge connecting u and v is labeled as e. Ignoring the direction of edges will not hinder the analysis of the noise problems.}
	\label{fig_noise_problem}
\end{figure}

Taking the graph in Figure \ref{fig_noise_problem} as an example, let $h_u^{(0)}$ be the initial attribute of node $u$ and $h_u^{(k)}$ be the attribute after $k$ layers of the network. 

When using the prevailing embedding methods to embed edge $e$, the embedding vectors $h_u^{(k)}$ and $h_v^{(k)}$ of nodes $u$ and $v$, respectively, are initially computed according to Equations \eqref{eq_common_node_embedding}.

\begin{align}
	h_u^{(k)} = \sigma(\text{AGGR}_n(h_u^{(k-1)}, M_i(u) \{n | n \in N_i(u), i \in [1,k]\} )) \nonumber \\
	h_v^{(k)} = \sigma(\text{AGGR}_n(h_v^{(k-1)}, M_i(v) \{n | n \in N_i(v), i \in [1,k]\} )),
	\label{eq_common_node_embedding}
\end{align}
where $\sigma$ represents the activation function used in the calculation of node embeddings, $\text{AGGR}_n$ represents the aggregation method, $N_i(u)$ denotes the $i$-th order neighbor nodes of $u$, and $M_i(u)$ stands for the mask matrix concerning the $i$-th order neighbor nodes.

Common graph learning methods, such as DGI \cite{DGI} and GDN \cite{GDN}, employ distinct $\sigma$, $\text{AGGR}_n$, and $M$ to adapt to various scenarios. Existing methods mostly compute the representation vector of edge $e$ based on the node embeddings, as shown in Equation \eqref{eq_common_edge_embedding}.

\begin{align}
	h_e^{(k)} &= f(h_u^{(k)}, h_v^{(k)}) \nonumber \\
	&= \sigma' (\text{AGGR}_e(h_u^{(0)},h_v^{(0)},M_i'\{h_n^{(0)}|n\in N(u) \cup N(v)\})),
	\label{eq_common_edge_embedding}
\end{align}
where $\sigma'$ and ${AGGR}_e$ denote abstract and general functions, while $M_i'$ represents the summarizing mask matrix.

Neighbor nodes of $u$ and $v$ that are not adjacent to $e$ do not exert direct influence on it. However, in Equation \eqref{eq_common_edge_embedding}, particularly when $i \geq 2$, high-order neighbor nodes play a substantial role in the computation of the embedding vector, notably in the portion denoted by $M_i'\{h_n^{(0)}|n\in N(u) \cup N(v)\}$. Furthermore, with an increase in $k$, i.e., the number of network layers, the edge embedding vector also incorporates information from nodes that are progressively more distant. These factors introduce noise to the embedding vector of $e$.

Several existing methods do not explicitly embed the edges, but they still suffer from the noise problem. Taking the highly representative UGED \cite{UGED} as an example, it relies on the bag-of-words model and samples the first-order neighbor nodes for each node, thereby computing the anomaly score for edges according to Equation \eqref{eq_UGED_general}.

\begin{align}
	e_{score} = \text{mean}(1-P(u|v,N_k(v)), 1-P(v|u,N_k(u))),
	\label{eq_UGED_general}
\end{align}
where $k$ is the number of samples.

This method aligns well with the idea of detecting anomalous edges based on node embeddings. However, it will introduce information from nodes $u_1$ and $v_1$ that are unrelated to $e$.

\begin{align}
	e_{score} = 1 - \text{mean}(\text{softmax}(\text{fc}(\text{mean}(u,N_k(u)))) + \nonumber \\
	\text{softmax}(\text{fc}(\text{mean}(v,N_k(v))))),
	\label{eq_UGED_specific}
\end{align}
where FC denotes a fully connected layer.

When the value of $k$ is small, the learning mechanism proposed by UGED faces challenges in capturing valuable information. Conversely, when the value of $k$ is large, it introduces a significant proportion of features from $u_1$ and $v_1$, thereby influencing the representation of the edge. Consequently, $N_k(u)$ and $N_k(v)$ introduce noise to the edge detection.

Similarly, in probability-based detection methods like AANE \cite{AANE}, the presence probability of edge $e$ is computed by evaluating the local structure surrounding it, given by $P_{uv}=sigmoid(uv)$. An indicator function denoted as $f(u,v)=P_{threshold}-P_{uv}$ is employed as an anomaly detection signal. Here, $P_{threshold}$ is computed by Equation \eqref{eq_AANE_P_threshold}.

\begin{align}
	P_{threshold} = \text{mean}(P_{un})-\mu \cdot \text{std}(P_{un}), n\in N(u),
	\label{eq_AANE_P_threshold}
\end{align}
where std() represents the standard deviation calculation function.

Let $k=\text{card}(N(u))$. The indicator function $f(u,v)$ can be computed using Equation \eqref{eq_AANE_indicator_function}.

\begin{align}
	f(u,v) =  & -\frac{\mu}{(k-1)^{\frac{1}{2}}}(\sum (P_{un'} - L - K)^2 + ((k-1)L-K)^2)^{\frac{1}{2}} \nonumber \\
	& + K - (k-1)L, n' \in N(u)-\{v\},
	\label{eq_AANE_indicator_function}
\end{align}
where $K=\frac{\sum P_{un'}}{k}$ and $L=\frac{P_{uv}}{k}$.

It is evident that $L$ is dependent on both $u$ and $v$, representing the nodes associated with edge $e$. On the other hand, the computation of $K$ requires sampling from the neighbors of node $u$, and various sampling methods can introduce information from nodes $u_1$ and even $u_2$, thereby introducing noise into the edge detection.

The remaining graph-based frameworks also face similar issues. With an increase in the number of layers, more irrelevant high-order neighbor nodes are introduced, leading to greater noise in edge representations.

In fact, when comparing edges $e$ and $e'$, it is meaningful to consider the attribute of node $u_1$ because the difference between $e$ and $e'$ predominantly hinges on the contrast between node pairs $(u, u_1)$ and $(u, v)$. Therefore, for nodes $u_1$, $v_1$, and others that are not directly linked to $e$, a more effective approach is to employ them as weight parameters in the explicit embedding of edges. This approach deviates from directly incorporating the node attributes into edge embeddings, thereby offering a promising avenue for addressing the noise problem.

\section{The Detection Framework}
\label{sec_framework}

\begin{figure*}[t]
	\centering
	\includegraphics[width=\linewidth]{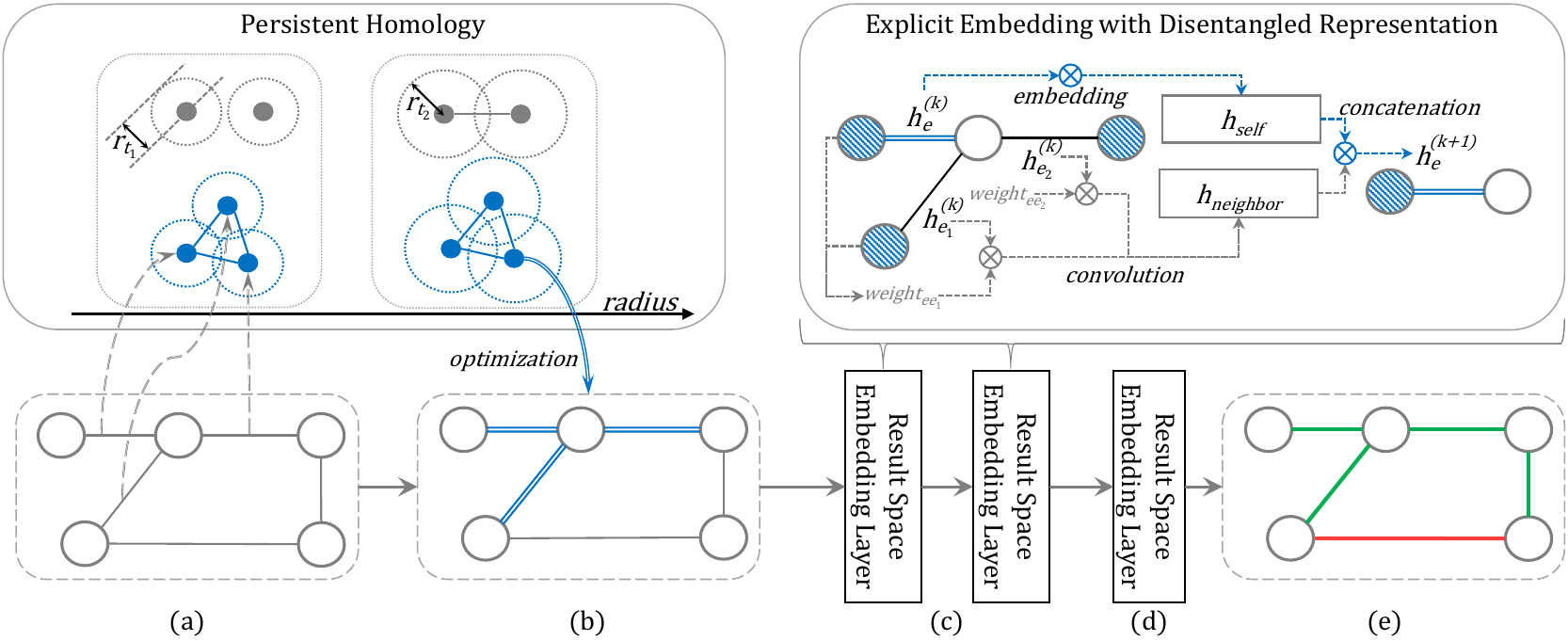}
	
	\caption{Overview of PhoGAD: (a) The graph constructed from behavioral data; (b) Optimization of edge attributes via persistent homology, where edges marked in blue with double lines correspond to enduring topological structures, and $r_\bullet$ represents the radius employed in persistent homology; (c) Explicit edge embedding with neighbor weights and disentangled representation, where nodes involved in weight calculation are shaded in blue, $weight_{e e_i}$ represents the weight between edge $e$ and $e_i$, and $h_e^{(k)}$ represents the embedding of edge $e$ after the $k$-th iteration; (d) Direct mapping of attributes to the output space after two layers of edge embedding; (e) Detection results, where edges highlighted in red are anomalous edges, indicating anomalous behaviors such as network intrusion or spam emails.}
	
	\Description{Overview of PhoGAD. The figure illustrates five key components of PhoGAD, as follows: (a) Constructing a graph based on behavioral data, where the edges are undirected to ensure the generalization; (b) Performing persistent homology to optimize edge attributes; (c) Displaying edge embeddings with neighbor weights and separation representation. Here, $\beta_{e e_i}$ represents the weight between edge $e$ and $e_i$, and $h_e^{(k)}$ represents the embedding of edge $e$ after the $k$-th iteration; (d) Directly mapping attributes to the result space after two layers of edge embeddings; (e) Detection results.}
	\label{fig_PhoGAD_framework}
\end{figure*}

We propose PhoGAD, a novel framework for anomaly behavior detection. To ensure the generalization to common behavior detection tasks, PhoGAD employs an undirected graph to model the raw data. It then utilizes persistent homology for edge attribute optimization. Subsequently, PhoGAD introduces adjacency edge weights to explicitly embed the edges, incorporating representation disentanglement in this process, ultimately achieving anomaly detection at the output layer. The overview of PhoGAD is illustrated in Figure \ref{fig_PhoGAD_framework}.

\subsection{Graph Construction}

Initially, a graph is constructed to represent the data and establish a formalized objective for detection.

\subsubsection{Nodes}

The nodes in the graph represent network entities, such as IP addresses, ports, and email accounts. These nodes possess attributes that encapsulate the features of these entities.

\subsubsection{Edges}

The edges in the graph represent specific network behaviors, such as intrusion and spam emails. The presence of an edge denotes the occurrence of a particular behavior, and the edge attributes correspond to the characteristics of that behavior, such as access features. Using this representation, the presence of anomalous edges in the graph directly indicates anomalous behaviors, making the detection of anomalous edges the primary objective.

In the scenarios mentioned in Section \ref{sec_anomaly_behaviors}, spam detection stands out as a specific case. Certain studies have employed complex models to extract textual features from emails \cite{spam_lstm, spam_transformer}. However, these methods frequently encounter challenges in capturing the interrelations between emails and demand domain-specific knowledge. The proposed PhoGAD, on the other hand, exclusively relies on fundamental word frequency as attributes. This accentuates the strengths of the graph and circumvents the necessity for domain knowledge.

\subsection{Persistent Homology Optimization}

Graph-based detection methods incorporate behaviors as nodes or edges, leveraging the interrelations between the behaviors. However, they do not directly exploit the inherent relationship between behaviors in the original feature space. Moreover, the relatively scattered attributes of normal edges can interfere with the detection results. To address this issue, we employ the topology analysis of persistent homology to optimize the attributes of edges. We extract the persistent topological structures and integrate the associated edge attributes to enhance specificity.

Persistent homology is a computational and encoding method tailored for the analysis of topological features across nested families of simplicial complexes and topological spaces. It empowers us to capture the complete view of the data without dimensionality reduction, thereby facilitating the accurate exploration of structural features \cite{tda_1, tda_2}. To capture the widely persistent structures, we extract the Vietoris-Rips complex denoted as $VR(x,\varepsilon)$, as depicted in Equation \eqref{eq_tda_vr}.

\begin{align}
	VR(x,\varepsilon)=\{<h_{e_0},h_{e_1},\dots,h_{e_k}>|B(h_{e_i},\varepsilon) \cap B(h_{e_j},\varepsilon) \neq 0\},
	\label{eq_tda_vr}
\end{align}
where $h_{e_i}$ represents the attribute of edge $e_i$, and $B(h_{e_i},\varepsilon)$ denotes the closed sphere centered at $h_{e_i}$ with a radius of $\varepsilon$.

As the radius $\varepsilon$ increases, the persistence of the topological structure formed by edge attributes can be plotted. Topological structures constructed by VR complexes across wider ranges of radii are considered as persistent non-anomalous patterns. Consequently, optimization of the attributes affiliated with the edges becomes feasible, as depicted in Equation \eqref{eq_tda_aggr}.

\begin{align}
	h_{e_i}'=\alpha h_{e_i} + (1-\alpha) \frac{1}{N}\sum_{e_j\in VR_{edges}} h_{e_j},
	\label{eq_tda_aggr}
\end{align}
where $VR_{edges}$ represents the set of edges contained in the persistent complex, and $h_{e_i}'$ denotes the updated attributes of edge $e_i$. 

This computation enables us to extract attributes from other edges present within the topological structures. Consequently, it introduces comprehensive discriminative insights on a global scale, thus overcoming the drawback of Euclidean distance-based representation methods that tend to disregard the inherent data structure. Furthermore, due to the insensitivity of topological analysis to data transformations, our approach remains effective across varying data scales, thereby enhancing its generalization.

\subsection{Detection via Explicit Edge Embedding}

Following the topological analysis,  we perform explicit embedding of edges and complete the detection. Within this process, we introduce the disentangled representation tailored for anomalous edge detection to address the noise problem. Additionally, we incorporate adjacency edge weights to handle the local heterophily of the graph, as demonstrated in Section~\ref{sec_problem_formulation}.

\subsubsection{Disentangled Representation}

Disentangled representation \cite{disentangled_representation} entails disentangling the representation of the current edge from the representations of its neighbor edges during convolution. To elaborate, for a given edge $e$, its current embedding $h_e$ is used to compute the embedding vector for the next round based on Equation \eqref{eq_disentangled_representation_general}.

\begin{align}
	h_e^{(k+1)} = \text{COMBINE}(h_e^{(k)},\text{AGGR}(\{ h_{e'}^{(k)} | e' \in N(e) \})),
	\label{eq_disentangled_representation_general}
\end{align}
where $N(e)$ represents the set of edges adjacent to $e$.

The initial attributes of edges describe the global features in the feature space, while the attributes of the neighbor edges capture the local features of the graph. Through convolution, these two sets of features can be amalgamated for accurate anomaly detection.

\subsubsection{Adjacent Edge Weights}

To ensure effective filtering of adjacent edges, we introduce the convolution parameter penalty based on node similarity, which is denoted by $\beta$ in Equations \eqref{eq_disentangled_representation_specific} and \eqref{eq_beta}. This allows the representations of edges connecting similar types of nodes to fuse more quickly, thereby facilitating the identification of localized anomalies.

\begin{align}
	h_e^{(k+1)} = & \sigma (W^{(k)} \cdot \nonumber \\
	& concat(f(h_e^{(k)}), \text{AGGR}(\{ \beta_{e,e'} h_{e'}^{(k)} | e' \in N(e) \}) )),
	\label{eq_disentangled_representation_specific}
\end{align}

\begin{align}
	\beta_{e,e'}  = \frac{{n_u \cdot n_v}}{{\lVert n_u \rVert \cdot \lVert n_v \rVert}},
	\label{eq_beta}
\end{align}
where $W^{(k)}$ represents trainable parameters, $\beta_{ee'}$ denotes the cosine similarity weight during the aggregation of edges $e$ and $e'$, $u$ and $v$ denote nodes that are adjacent and mutually second-order neighbors based on $e$ and $e'$, and $n_u$ and $n_v$ are their attributes. The activation function and dropout function are denoted by $\varepsilon$. This approach ensures the generalization of PhoGAD across various scenarios.

\subsubsection{The Detection}

Due to the typically low proportion of anomalous behaviors in anomaly network behavior detection, we adapt the focal loss from the domain of object detection \cite{focal_loss}. The computation of the loss using Equation \eqref{eq_focal_loss}.

\begin{align}
	Loss = (y-1)(1-\delta)p^\gamma \log(1-p) - y \delta (1-p)^\gamma \log p,
	\label{eq_focal_loss}
\end{align}
where $y$ represents the sample label and $p$ denotes the detection value.

The obtained edge embeddings exhibit discriminative properties, so we only need to map them to a two-dimensional detection result space through a linear layer. This approach realizes the anomaly detection for edges, thereby uncovering anomalous behaviors present within the raw data.

\section{Experiments and Results}
\label{sec_experiments}

\subsection{Experiment Setup}

\subsubsection{Datasets}
\label{sec_datesets}
To evaluate the detection capabilities of PhoGAD across various scenarios, we employed publicly available datasets from three distinct domains: network intrusion detection, anonymous traffic detection, and spam detection.

The UNSW-NB15 dataset (UNSW) \cite{data_unsw_1} is employed for the network intrusion detection scenario. It encompasses an extensive collection of network traffic samples, encompassing both normal network traffic and diverse malicious behaviors, such as Denial of Service attacks, malware incidents, and scans. Additionally, the ISCXTor2016 dataset (TOR) \cite{data_tor} is utilized for the anonymous traffic detection scenario. It comprises a diverse set of traffic samples originating from the Tor network, all of which have been labeled. Lastly, the SpamAssassin dataset (SPAM) \cite{data_SpamAssassin} is employed for the spam detection scenario. It includes both legitimate emails and spam samples, offering coverage across various types of emails.

\subsubsection{Baselines}

We have employed several baselines to compare and illustrate the detection performance of PhoGAD.

CSAWD \cite{CSAWD} utilizes disparities within node embeddings to detect anomalous edges. Besides, node embeddings computed by OCGNN \cite{OCGNN} are directly mapped into edge detection results through neural networks. Both of these frameworks are adopted as representatives for implicit representation-based detection.
GNN-NIDS \cite{GNN-NIDS} portrays entities and behaviors as nodes, employing node detection methods to detect anomalous behaviors. Anomal-E \cite{Anomal-E}, based on contrastive learning, utilizes adjusted GraphSage \cite{GraphSage} for edge embedding and subsequent detection. These two frameworks are adopted as representatives for explicit representation-based detection.
Additionally, AnomalyDAE \cite{AnomalyDAE}, recognized as the top-performing framework among graph-based autoencoder detection frameworks, is adopted as the representative for detection based on reconstruction error.

Among the baselines, Anomal-E stands out as the SOTA framework for detecting anomalous behaviors through anomalous edge detection. AnomalyDAE has demonstrated promising performance in detecting anomalies in specific scenarios.

\subsubsection{Metrics and Parameter Settings}

We assess specific detection performance using accuracy, precision, and recall. The F1 score is used to assess the overall performance.

The experiments are implemented using the PyTorch and MindSpore\footnote{https://www.mindspore.cn/} frameworks.
Thanks to the considerable accuracy attained via topology-based structural mining, the value of alpha in Equation \ref{eq_tda_aggr} is set to 0.7, thereby enhancing the differentiation between normal and anomalous edges. In loss function \ref{eq_focal_loss}, both $\delta$ and $\gamma$ are set to 2 to accommodate scenarios where the anomaly proportion is extremely low. Within the framework, the first layer's dimension is established at one-fourth of the input dimension, the second layer's dimension is set to half of the first layer's dimension, and the output dimension is configured as 2. Since the optimizer is not the focus, we adopt the widely used Adam optimizer.

Given the elevated anomaly proportion within the datasets, which diverges from the typical distribution observed in real-world scenarios, we perform random sampling on the original datasets to ensure a maximum anomaly proportion of 10\%.

\subsection{Persistent Homology Analysis Results}

\begin{figure}[h]
	\centering
	
	\subfigure[Persistence Diagram of UNSW]{
		\label{unsw_persistence_diagram}
		\includegraphics[width=0.48\linewidth]{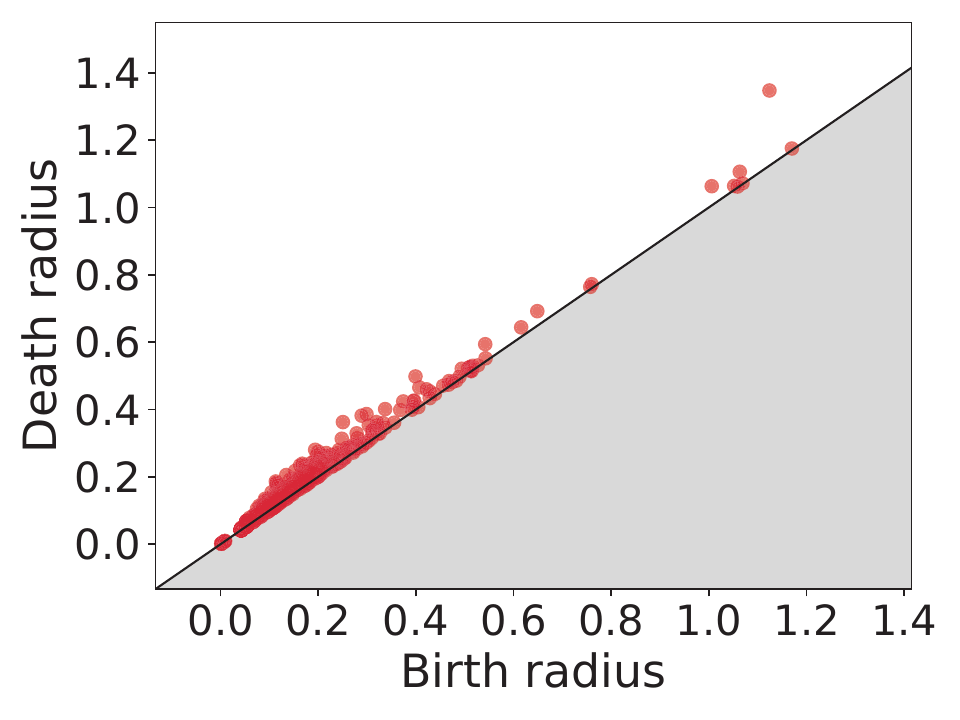}}
	\hfill
	\subfigure[Barcode Diagram of UNSW]{
		\label{unsw_barcode_diagram}
		\includegraphics[width=0.48\linewidth]{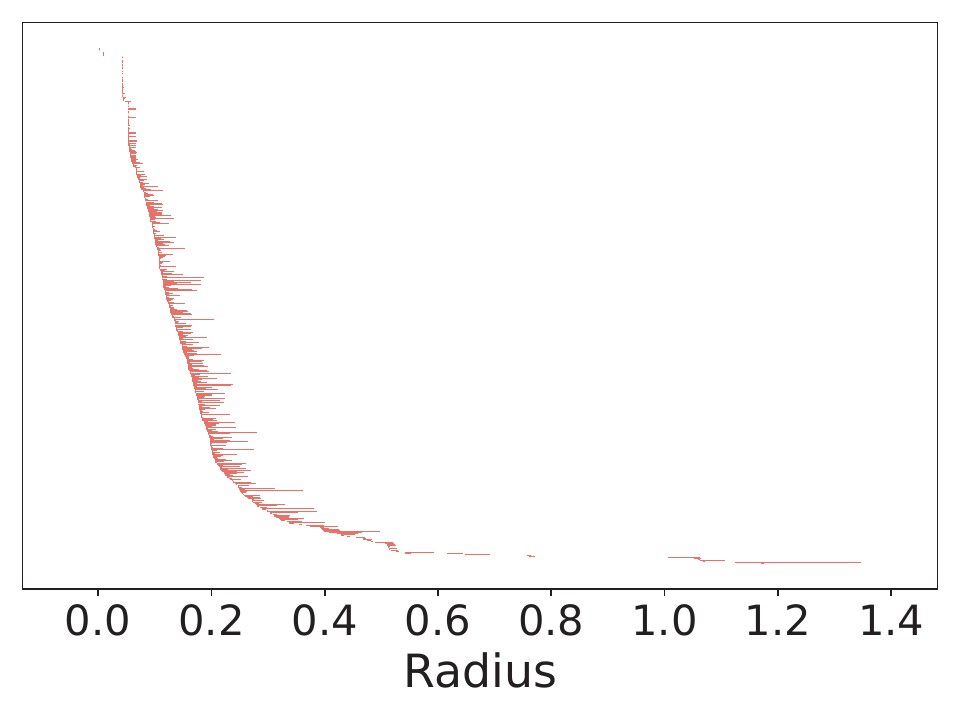}}
	
	\subfigure[Persistence Diagram of TOR]{
		\label{tor_persistence_diagram}
		\includegraphics[width=0.48\linewidth]{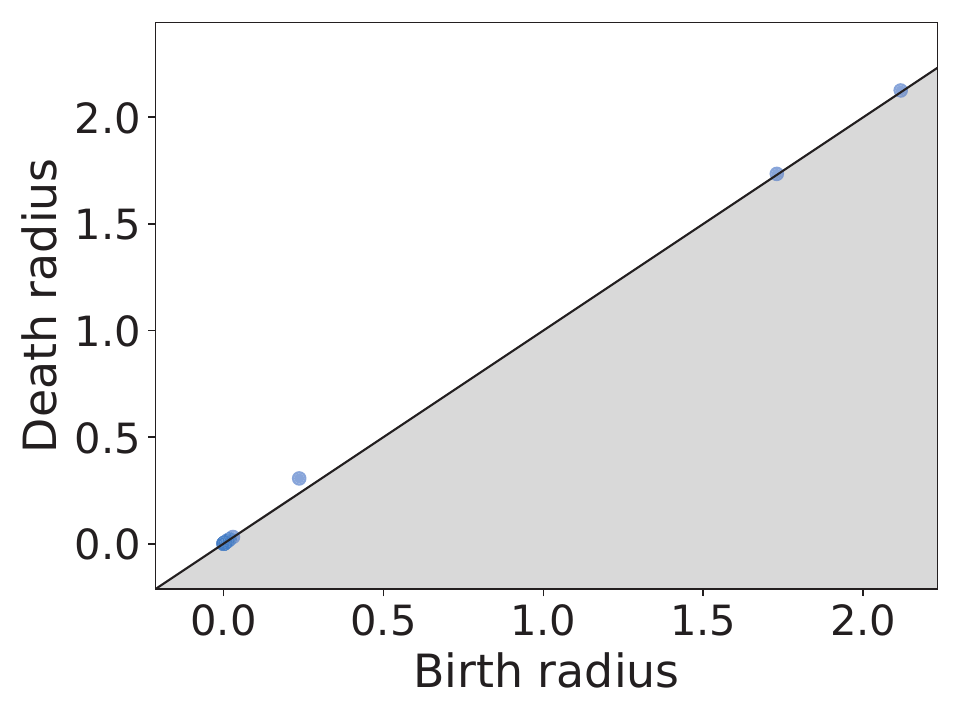}}
	\hfill
	\subfigure[Barcode Diagram of TOR]{
		\label{tor_barcode_diagram}
		\includegraphics[width=0.48\linewidth]{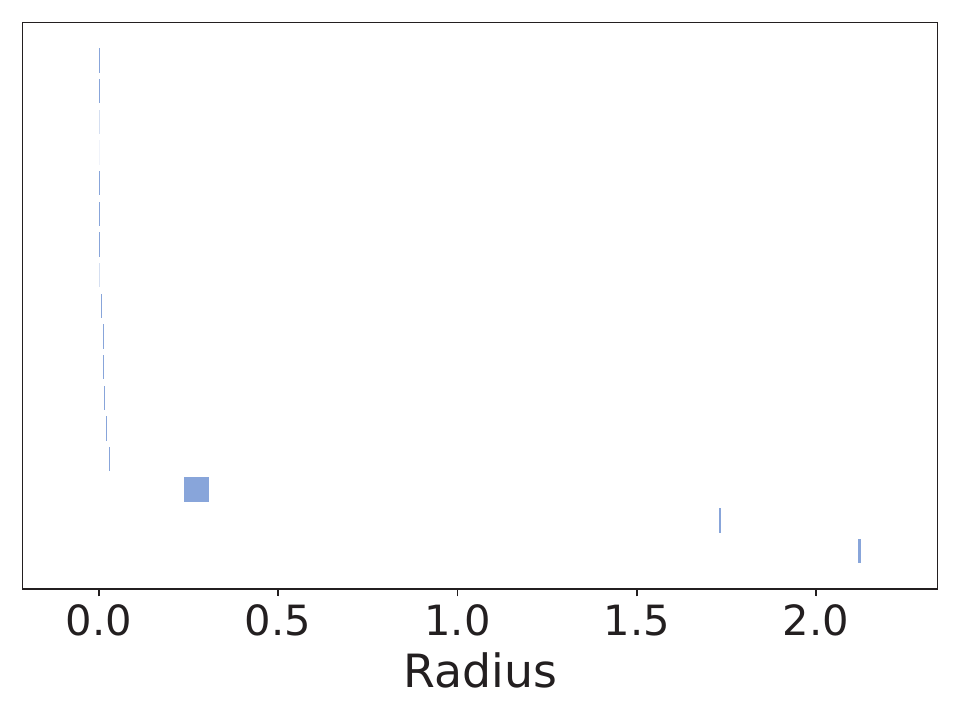}}
	
	\subfigure[Persistence Diagram of SPAM]{
		\label{spam_persistence_diagram}
		\includegraphics[width=0.48\linewidth]{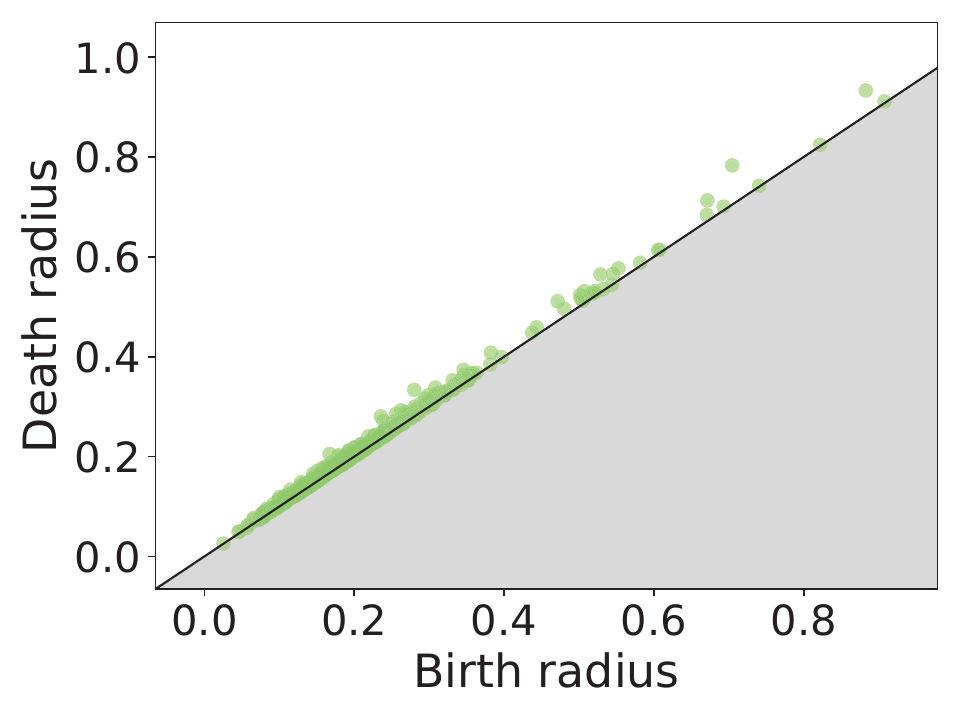}}
	\hfill
	\subfigure[Barcode Diagram of SPAM]{
		\label{spam_barcode_diagram}
		\includegraphics[width=0.48\linewidth]{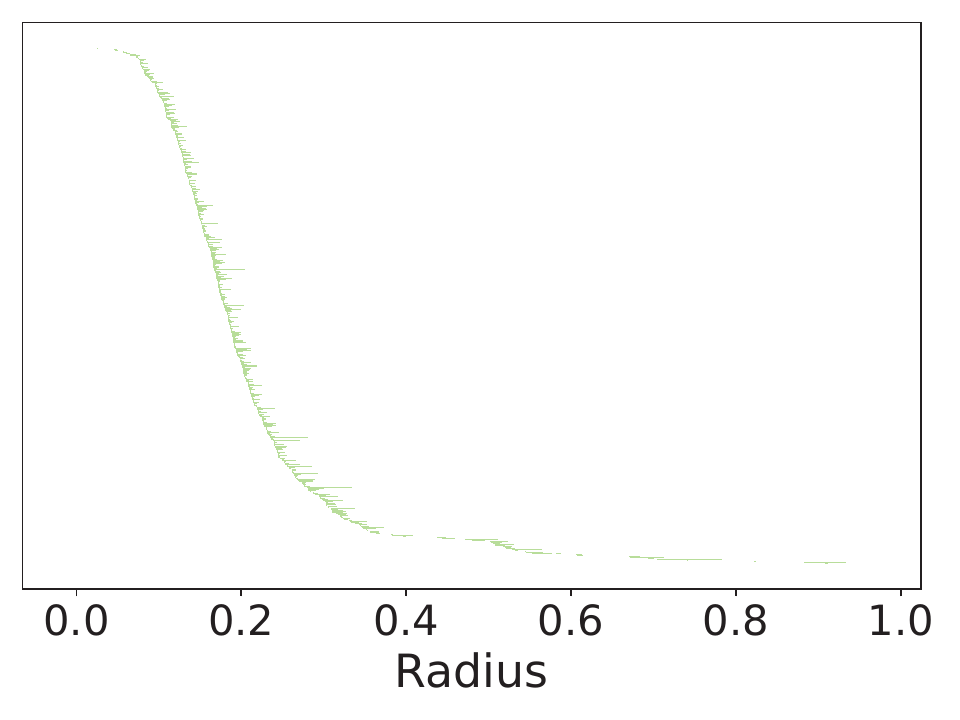}}
	
	\caption{Results of persistent homology. (a), (c), and (e) are persistence diagrams, where the horizontal axis denotes the radius $\varepsilon$ at which the corresponding topological structure appears, while the vertical axis denotes $\varepsilon$ at which it vanishes. (b), (d), and (f) are barcode diagrams, with the left and right endpoints of each bar indicating $\varepsilon$ at which the structure appears and vanishes.}

	\Description{Results of persistent homology, containing six subfigures. (a), (c), and (e) are persistence diagrams, while (b), (d), and (f) are barcode diagrams. Figures (a) and (b) illustrate the results of persistent homology on UNSW, whereas (e) and (f) display that on SPAM. Both datasets exhibit a significant presence of non-persistent topological structures alongside several persistent ones. Figures (c) and (d) showcase the results of persistent homology on TOR, revealing only a few persistent 1-dimensional structures.}
	\label{fig_tda_results}
\end{figure}

Figure \ref{fig_tda_results} illustrates the results of topological analysis through persistent homology. Given that 0-dimensional simplices represent fundamental connected structures without more intricate topological features, we disregard them and focus solely on 1-dimensional simplices, which correspond to 1-dimensional topological holes.

A point positioned farther away from the diagonal line in the persistence figure, or a longer bar in the barcode diagram, indicates that the corresponding topological structure persists for a longer duration.
Due to the normalization of all feature fields, the points exhibit lesser deviation from the diagonal line in terms of visualization. Nonetheless, it remains evident that certain behavior features have formed persistent structures, although the majority of structures do not persist.

Figures \ref{tor_persistence_diagram} and \ref{tor_barcode_diagram} exhibit a discernible contrast compared to the others. This is because of the relatively uniform distribution of edge features in TOR, which does not generate a substantial quantity of 1-dimensional simplices. As a result, persistent homology identifies a smaller number of topological structures, but relatively persistent structures still manifest within this subset.

Table \ref{table_tda_percentage} further presents the proportion of normal and anomalous edges within the set corresponding to the mined persistent topological structures.

\begin{table}[h]
	\caption{Proportion of Edges in Mined Persistent Structures}
	\label{table_tda_percentage}
	\resizebox{\linewidth}{!}{
		\begin{tabular}{lcc}
			\toprule
			\textbf{Dataset} & \textbf{Proportion of Anomaly Edges} & \textbf{Proportion of Normal Edges}\\
			\midrule
			UNSW & 0.0294 & 0.9706 \\
			TOR & 0.0328 & 0.9672 \\
			SPAM & 0.0571 & 0.9429 \\
			\bottomrule
		\end{tabular}
	}
\end{table}

It is evident that the majority of edges corresponding to the discovered structures are normal, indicating the prolonged persistence of normal patterns. This observation aligns with the assumption of anomaly detection. Therefore, the optimization has the potential to heighten the resemblance among normal edges, thereby alleviating the issue of the ambiguous behavioral boundary.

\subsection{Anomaly Behavior Detection Results}

\subsubsection{Overall Detection Results}
Table \ref{table_overall_detection_results} presents the detection results of PhoGAD and the baselines on each dataset, where the anomaly proportion has been reduced to 10\%.

\begin{table*}
	\caption{Overall Detection Results of all Frameworks}
	\label{table_overall_detection_results}
	\resizebox{\linewidth}{!}{
		\begin{tabular}{lcccccccccccc}
			\toprule
			& \multicolumn{4}{@{}c@{}}{\textbf{Results on UNSW}} & \multicolumn{4}{@{}c@{}}{\textbf{Results on TOR}} & \multicolumn{4}{@{}c@{}}{\textbf{Results on SPAM}} \\
			\cmidrule(lr){2-5}
			\cmidrule(lr){6-9}
			\cmidrule(lr){10-13}
			\textbf{Framework} & accuracy & recall & precision & F1 & accuracy & recall & precision & F1 & accuracy & recall & precision & F1 \\
			\midrule
			CSAWD \cite{CSAWD} & 0.9736 & 0.8548 & 0.8776 & 0.8660 & 0.9652 & 0.7926 & 0.8492 & 0.8199 & 0.9558 & 0.7583 & 0.7913 & 0.7744 \\
			OCGNN \cite{OCGNN} & 0.9848 & 0.9370 & 0.9134 & 0.9250 & 0.9833 & 0.8926 & 0.9377 & 0.9146 & 0.9629 & 0.7708 & 0.8447 & 0.8061 \\
			GNN-NIDS \cite{GNN-NIDS} & 0.9756 & 0.8593 & 0.8923 & 0.8755 & 0.9763 & 0.8963 & 0.8705 & 0.8832 & 0.9610 & 0.7767 & 0.8233 & 0.7993 \\
			Anomal-E \cite{Anomal-E} & 0.9972 & 0.9852 & 0.9866 & 0.9859 & 0.9901 & 0.9274 & \textbf{0.9728} & 0.9496 & 0.9658 & 0.7917 & 0.8559 & 0.8225 \\
			AnomalyDAE \cite{AnomalyDAE} & 0.9867 & 0.9704 & 0.9034 & 0.9357 & 0.9793 & 0.9037 & 0.8905 & 0.8971 & 0.9804 & 0.9208 & 0.8876 & 0.9039 \\
			PhoGAD (Ours) & \textbf{0.9990} & \textbf{0.9959} & \textbf{0.9941} & \textbf{0.9950} & \textbf{0.9907} & \textbf{0.9333} & 0.9722 & \textbf{0.9524} & \textbf{0.9875} & \textbf{0.9325} & \textbf{0.9419} & \textbf{0.9372} \\	
			\bottomrule
		\end{tabular}
	}
\end{table*}

The initial observation is that PhoGAD attained the most favorable overall detection performance across nearly all datasets and metrics. Notably, on the SPAM dataset, where we only employ word frequencies as initial edge attributes, the detection performance obtained is independent of domain knowledge. These results demonstrate that the disentangled representation and adjacency edge weight configuration effectively address the noise problem and the local heterophily of the graph.

The exception to this trend lies in TOR dataset, where PhoGAD and Anomal-E demonstrated minimal divergence, with PhoGAD exhibiting a marginally lower precision. This can be attributed to the limited number of persistent topological structures unveiled by persistent homology, as shown in Figures \ref{tor_persistence_diagram} and \ref{tor_barcode_diagram}. Consequently, normal features were not adequately emphasized, leading to an increased occurrence of false positives and a decrease in precision.

Among the remaining baselines, Anomal-E outperforms OCGNN, indicating the superiority of explicit representation over implicit representation, as well as the effectiveness of direct edge detection. On the other hand, CSAWD and GNN-NIDS exhibit relatively poor performance, primarily due to the significant amount of noise introduced in their approaches, resulting in a comparatively severe noise problem. Worth noting is the behavior of AnomalyDAE, its recall consistently surpasses its precision, suggesting a lower rate of false negatives. However, the reduced precision adversely impacts the comprehensive metric, the F1 score.

\subsubsection{Results of Extremely Low Anomaly Proportions}

In real-world scenarios, anomalies such as intrusions and anonymous traffic typically have extremely low proportions compared to anomalies like spam emails \cite{proportion_is_low_1}. Therefore, we conducted additional experiments by reducing the anomaly proportions to evaluate the efficacy of each framework under conditions that more closely resemble real-world scenarios. Table \ref{table_low_proportion_5_results} presents the results when the proportion of anomalies is reduced to 5\%.

\begin{table}[h]
	\caption{Detection Results at 5\% Anomaly Proportion}
	\label{table_low_proportion_5_results}
	\resizebox{\linewidth}{!}{
		\begin{tabular}{lcccccccc}
			\toprule
			& \multicolumn{4}{@{}c@{}}{\textbf{Results on UNSW}} & \multicolumn{4}{@{}c@{}}{\textbf{Results on TOR}} \\
			\cmidrule(lr){2-5}
			\cmidrule(lr){6-9}
			\textbf{Framework} & accuracy & recall & precision & F1 & accuracy & recall & precision & F1\\
			\midrule
			CSAWD \cite{CSAWD} & 0.9644 & 0.6889 & 0.6327 & 0.6596 & 0.9607 & 0.5259 & 0.6283 & 0.5726 \\
			OCGNN \cite{OCGNN} & 0.9896 & 0.8578 & 0.9294 & 0.8922 & 0.9530 & 0.7030 & 0.8068 & 0.7513 \\
			GNN-NIDS \cite{GNN-NIDS} & 0.9741 & 0.7259 & 0.7481 & 0.7368 & 0.9804 & 0.7556 & 0.8361 & 0.7938 \\
			Anomal-E \cite{Anomal-E} & 0.9950 & 0.9407 & 0.9592 & 0.9499 & 0.9874 & 0.8667 & 0.8797 & 0.8732 \\
			AnomalyDAE \cite{AnomalyDAE} & 0.9822 & 0.9556 & 0.7544 & 0.8432 & 0.9839 & 0.9608 & 0.7903 & 0.8672 \\
			PhoGAD (Ours) & \textbf{0.9989} & \textbf{0.9914} & \textbf{0.9971} & \textbf{0.9942} & \textbf{0.9878} & \textbf{0.9533} & \textbf{0.9303} & \textbf{0.9417} \\
			\bottomrule
		\end{tabular}
	}
\end{table}

As observed, the proposed PhoGAD consistently sustains good performance even when confronted with extremely low anomaly proportion, despite a slight decrease in certain metrics. Particularly, in contrast to the results presented in Table \ref{table_overall_detection_results}, PhoGAD outperforms Anomal-E across all metrics.

On the other hand, the performance of implicit embedding frameworks, as exemplified by OCGNN, shows a notable decline. It is worth noting that AnomalyDAE, rooted in autoencoders, is less affected by the decrease in anomaly proportions compared to other baselines, maintaining a higher recall.

We further decreased the anomaly proportion to 3\%, and the results are shown in Table \ref{table_low_proportion_3_results}.

\begin{table}[h]
	\caption{Detection Results at 3\% Anomaly Proportion}
	\label{table_low_proportion_3_results}
	\resizebox{\linewidth}{!}{
		\begin{tabular}{lcccccccc}
			\toprule
			& \multicolumn{4}{@{}c@{}}{\textbf{Results on UNSW}} & \multicolumn{4}{@{}c@{}}{\textbf{Results on TOR}} \\
			\cmidrule(lr){2-5}
			\cmidrule(lr){6-9}
			\textbf{Framework} & accuracy & recall & precision & F1 & accuracy & recall & precision & F1\\
			\midrule
			CSAWD \cite{CSAWD} & 0.9652 & 0.4321 & 0.4217 & 0.4268 & 0.9653 & 0.3086 & 0.3994 & 0.3482 \\
			OCGNN \cite{OCGNN} & 0.9874 & 0.7037 & 0.8507 & 0.7702 & 0.9833 & 0.6420 & 0.7647 & 0.6980 \\
			GNN-NIDS \cite{GNN-NIDS} & 0.9788 & 0.7630 & 0.6192 & 0.6836 & 0.9763 & 0.6543 & 0.5955 & 0.6235 \\
			Anomal-E \cite{Anomal-E} & 0.9945 & 0.8716 & 0.9413 & 0.9051 & 0.9881 & 0.7407 & 0.8451 & 0.7895 \\
			AnomalyDAE \cite{AnomalyDAE} & 0.9870 & 0.9506 & 0.7130 & 0.8148 & 0.9622 & \textbf{0.9753} & 0.4413 & 0.6077 \\
			PhoGAD (Ours) & \textbf{0.9982} & \textbf{0.9677} & \textbf{0.9745} & \textbf{0.9711} & \textbf{0.9929} & 0.8519 & \textbf{0.9055} & \textbf{0.8779} \\
			\bottomrule
		\end{tabular}
	}
\end{table}

Implicit embedding frameworks exhibit a sharp decline in performance. In contrast, PhoGAD maintains relatively high performance, second only to AnomalyDAE in recall on the UNSW dataset, but significantly leading in all other metrics. This suggests that reconstruction-based frameworks sustain a low false negative rate under extremely low anomaly proportions but tend to generate a notable volume of false positives.

These results further demonstrate that the proposed PhoGAD can effectively isolate the interplay between anomalous and normal samples, enabling it to uphold robust performance even when confronted with a low anomaly proportion.

\subsection{Ablation Studies}

To evaluate the practical effect of the mechanisms employed by PhoGAD, we conduct ablation experiments within the intrusion detection scenario. Table \ref{table_ablation_results} presents the results. To obtain meaningful detection results, the anomaly proportion is set to 5\%.

\begin{table}[h]
	\caption{Results of Ablation Experiments}
	\label{table_ablation_results}
	\resizebox{\linewidth}{!}{
		\begin{tabular}{lcccc}
			\toprule
			\textbf{Framework} & \textbf{accuracy} & \textbf{precision} & \textbf{recall} & \textbf{F1}\\
			\midrule
			Without disentangled representation & 0.9854 & 0.9429 & 0.9141 & 0.9283 \\
			Without neighbor weights & 0.9931 & 0.9686 & 0.9631 & 0.9658 \\
			Without persistent homology & 0.9854 & 0.9743 & 0.8903 & 0.9304 \\
			Without any novel mechanisms & 0.9771 & 0.9371 & 0.8497 & 0.8913 \\
			Complete PhoGAD & \textbf{0.9989} & \textbf{0.9914} & \textbf{0.9971} & \textbf{0.9942} \\
			\bottomrule
		\end{tabular}
	}
\end{table}

Evidently, the mechanisms collectively contribute to improving the overall detection performance, with persistent homology making a particularly significant contribution. Among the mechanisms, persistent homology places a greater emphasis on reducing the false positive rate, while adjacent edge weights and disentangled representation focus on reducing the false negative rate.

This phenomenon arises since persistent homology leads to a more concentrated distribution of attributes of normal edges. Conversely, the other two mechanisms primarily focus on segregating anomalous edges from normal ones, thereby protecting the distinctive attributes from being obscured, especially in scenarios where the anomaly proportion is low.

\section{Conclusion}
\label{sec_conclusion}

In this paper, we propose PhoGAD, a graph-based anomaly behavior detection framework. PhoGAD introduces persistent homology optimization, second-order node-based adjacency edge weights, and disentangled representation to address the challenges of ambiguous behavioral boundary, local heterophily of graphs, and the noise problem in real-world network environments.
PhoGAD has demonstrated superior detection performance compared to baselines in various scenarios. Furthermore, it exhibits adaptability to extremely low anomaly proportions. To the best of our knowledge, PhoGAD is the first framework that incorporates topological analysis and optimized explicit edge embeddings for network anomaly behavior detection. Looking ahead, our future research will delve into investigating more complex persistent structures, as well as refining the synergy between topological analysis and graph neural networks to further enhance learning and detection capabilities.

\begin{acks}

The authors of this paper were supported by the National Natural Science Foundation of China through grants No.62225202 and No.62202029. This work was also sponsored by CAAI-Huawei MindSpore Open Fund.

\end{acks}

\section*{Ethical Considerations}

PhoGAD, as an anomaly detection framework, inherently yields positive societal impacts within its intended scope. However, it is crucial to recognize potential negative consequences. For instance, there is a chance that certain instances of anonymous traffic may not be driven by malicious intent but rather aim to safeguard privacy. PhoGAD's application could, in such cases, misclassify such traffic as anomalies. Furthermore, in real-world scenarios, abrupt changes in behavioral patterns resulting from external factors, such as unanticipated download spikes linked to the imminent closure of a file-sharing service, may also be identified as anomalies by PhoGAD. Such misclassifications possess the potential to disrupt ordinary user activities.

To address these drawbacks, comprehensive considerations are necessary in both technical and application domains. Technically, PhoGAD should refrain from employing content analysis methods such as protocol parsing to safeguard privacy in cyberspace. Moreover, in practical situations, the implementation of a whitelist strategy proves effective in accommodating expected deviations in normal behavioral patterns. On the application front, the utilization of PhoGAD must align with compliant management practices, avoiding the indiscriminate expansion of the detection scope. Collectively, these strategies work synergistically to mitigate or even prevent potential adverse effects arising from the deployment of PhoGAD.

\bibliographystyle{ACM-Reference-Format}
\balance
\bibliography{reference}

\end{document}